\documentclass[sigconf]{acmart}

\AtBeginDocument{%
  \providecommand\BibTeX{{%
    \normalfont B\kern-0.5em{\scshape i\kern-0.25em b}\kern-0.8em\TeX}}}

\setcopyright{acmlicensed}
\copyrightyear{2024}
\acmYear{2024}
\acmDOI{XXXXXXX.XXXXXXX}

\acmConference[CIKM ’24]{In Proceedings of the 32nd ACM International Conference on Information and Knowledge Management}{October 21 – 25, 2024}{Boise, Idaho}
\acmISBN{978-1-4503-XXXX-X/18/06}

\usepackage{multirow} 
\usepackage{caption}

\begin{document}

\title{Uncertainty-Aware Crime Prediction With Spatial Temporal Multivariate Graph Neural Networks}


\author{Zepu Wang}
\affiliation{%
  \institution{University of Pennsylvania}
  \country{USA}}
\email{zepu@seas.upenn.edu}

\author{Xiaobo Ma}
\affiliation{%
  \institution{University of Arizona}
  \country{USA}}
\email{xiaoboma@arizona.edu}

\author{Huajie Yang}
\affiliation{%
  \institution{City University of Macau}
  \country{China}}
\email{yanghj@cityu.edu.mo}

\author{Weimin Lvu}
\affiliation{%
  \institution{Stony Brook University}
  \country{USA}}
\email{Weimin.lyu@stonybrook.edu}

\author{Peng Sun}
\authornote{Corresponding Authors.}
\affiliation{%
  \institution{Duke Kunshan University}
  \country{China}}
\email{peng.sun568@duke.edu}

\author{Sharath Chandra Guntuku}
\affiliation{%
  \institution{University of Pennsylvania}
  \country{USA}}
\email{sharathg@seas.upenn.edu}

\renewcommand{\shortauthors}{Wang et al.}

\begin{abstract}
Crime forecasting is a critical component of urban analysis and essential for stabilizing society today. Unlike other time series forecasting problems, crime incidents are sparse, particularly in small regions and within specific time periods. Traditional spatial-temporal deep learning models often struggle with this sparsity, as they typically cannot effectively handle the non-Gaussian nature of crime data, which is characterized by numerous zeros and over-dispersed patterns. To address these challenges, we introduce a novel approach termed Spatial Temporal Multivariate Zero-Inflated Negative Binomial Graph Neural Networks (STMGNN-ZINB). This framework leverages diffusion and convolution networks to analyze spatial, temporal, and multivariate correlations, enabling the parameterization of probabilistic distributions of crime incidents. By incorporating a Zero-Inflated Negative Binomial model, STMGNN-ZINB effectively manages the sparse nature of crime data, enhancing prediction accuracy and the precision of confidence intervals. Our evaluation on real-world datasets confirms that STMGNN-ZINB outperforms existing models, providing a more reliable tool for predicting and understanding crime dynamics.
\end{abstract}



\maketitle

\section{Introduction}
Accurate crime forecasting can significantly enhance police deployment strategies and infrastructure allocation, substantially improving urban safety~\cite{wang2022hagen, huang2018deepcrime, huang2019mist}. With advancements in deep learning~\cite{lin2023mmst,ke2020enhancing,liu2024enhancing,bautista2022autonomous,wang2022novel}, researchers are increasingly utilizing complex neural network architectures to model crime patterns. These include recurrent neural networks~\cite{huang2018deepcrime}, convolutional neural networks~\cite{safat2021empirical}, and graph neural networks~\cite{sun2021crimeforecaster, huang2019mist}. These models aim to capture both the spatial-temporal correlations of crime patterns and the interrelationships among different crime categories. Specifically, spatial-temporal graph neural networks are good at extracting spatial-temporal correlations from urban data~\cite{mo2024cross,ruan2024infostgcan,wang2022novel1}.

However, the deterministic models predominantly used in crime forecasting implicitly presuppose that the outputs follow a Gaussian distribution~\cite{zhuang2022uncertainty, almanie2015crime,huang2018deepcrime,he2023uncertainty,he2023survey}, significantly simplifying the variance structure~\cite{ding2017using,guo2014adaptive}. Despite their effectiveness, most existing spatial-temporal prediction methods face limitations when predicting urban crimes due to the sparsity of crime patterns. The data for each fine-grained urban region is extremely sparse~\cite{wu2020hierarchically,zhao2016multi}. A plethora of zero values in crime data indicates that it is not appropriate to assume that crime data patterns follow a Gaussian distribution. Research demonstrates the effectiveness of the Zero-Inflated Negative Binomial (ZINB) distribution in analyzing sparse travel demand data~\cite{zhuang2022uncertainty} and sparse traffic risk forecasting~\cite{gao2023spatiotemporal}. Additionally, uncertainty quantification with ZINB has been conducted for many urban and transportation tasks~\cite{jiang2023uncertainty, zhuang2023sauc}.

In this paper, we propose the Spatial Temporal Multivariate Zero-Inflated Negative Binomial Graph Neural Networks (STMGNN-ZINB)—a comprehensive framework designed for joint numeric prediction and uncertainty quantification of urban crime. Our main contributions are as follows:

\begin{itemize}
    \item We utilize the Zero-Inflated Negative Binomial (ZINB) distribution to model crime, effectively capturing zero-inflation and addressing data sparsity.
    \item We integrate ZINB with spatial-temporal multivariate graph neural networks, enabling precise quantification of the sparse and discrete uncertainty in crime data.
    \item We demonstrate the superiority of STMGNN-ZINB through experiments on two real-world crime datasets, comparing it against various other distribution assumptions and performance metrics.
\end{itemize}

The paper is organized as follows: Section~\ref{pre} defines the research question and develops the model. Section~\ref{exp} details the datasets used, the evaluation metrics, and presents the experimental results. Section~\ref{con} concludes the paper.

\section{Preliminaries}
\label{pre}

\subsection{Problem Definition}

Formally, a graph \(G\) is defined as an ordered pair \((V, E)\), where \(V\) represents the set of vertices (or nodes) and \(E\) represents the set of edges. Suppose the historical crime data is embedded in graph \(G\) with \(C\) multivariate crime features over \(T\) time intervals. The historical time series features can then be represented as \(X \in \mathbb{R}^{N \times T \times C}\).

The objective of this task is to learn a mapping function \(f\) that uses the historical crime data \(X\) and the graph structure \(G\) as inputs to forecast future crime data for \(Q\) time intervals. Our goal is not only to predict the expected values of future crime but also to estimate the confidence intervals for these predictions. Consequently, we denote our output as \(\hat{X} \in \mathbb{R}^{N \times Q \times C \times Z}\), where \(Z\) represents the parameters of the assumed distribution of crime data.

\subsection{Zero-Inflated Negative Binomial Distribution}

In our model, we assume that the distribution of crimes follows the Zero-Inflated Negative Binomial (ZINB) distribution~\cite{jiang2018sparse,zhuang2022uncertainty}. The probability mass function (PMF) of a random variable following the ZINB distribution is given by:

\begin{equation}
    P(Y=y) = 
\begin{cases} 
\pi + (1-\pi)(1-p)^r & \text{if } y = 0 \\
(1-\pi) \binom{y + r - 1}{y} p^y (1-p)^r & \text{if } y = 1,2,3,\dots
\end{cases}
\end{equation}

Here, \(\pi\) represents the probability of an extra zero, indicating zero inflation, which in this context, signifies the likelihood of no occurrences of a specific type of crime. The parameters \(r\) and \(p\) are the shape parameters of the traditional negative binomial distribution, where \(r\) affects the dispersion and \(p\) is the probability of success in each experiment.

\subsection{Diffusion Graph Convolution Networks (DGCNs)}

To capture spatial correlations within a predefined graph structure, we utilize Diffusion Graph Convolutional Networks (DGCNs). These models employ the concept of diffusion processes on graphs to effectively capture the spread of information across the graph’s topology.

At the heart of DGCNs is the diffusion convolution operation, which can be considered a generalization of the traditional convolutional operations adapted for graph data. The underlying principle is to simulate a diffusion process on the graph, allowing information to propagate from a node to its neighbors through multiple steps or layers. This dynamic can be mathematically articulated using the graph Laplacian and its exponentiations to represent various degrees of diffusion.

The diffusion convolution operation in a DGCN is mathematically defined as:

\[
H^{(l+1)} = \sigma\left(D^{-1}A H^{(l)}W^{(l)} + B^{(l)}H^{(l)}\right)
\]

where:
- \(H^{(l)}\) denotes the node features (or hidden states) at layer \(l\),
- \(A\) is the adjacency matrix of the graph, where \(A_{ij}\) represents the edge weight between nodes \(i\) and \(j\), with \(A_{ij} = 0\) if no edge exists,
- \(D\) is the diagonal degree matrix, with each diagonal element \(D_{ii}\) being the sum of the weights of all edges connected to node \(i\),
- \(W^{(l)}\) is the weight matrix for layer \(l\), which is learned during the training process,
- \(B^{(l)}\) is the bias term for layer \(l\),
- \(\sigma(\cdot)\) is a non-linear activation function, such as the Rectified Linear Unit (ReLU).

This formulation enables the network to effectively balance the influence of immediate neighbors (through \(D^{-1}A H^{(l)}\)) and the retention of the current node’s features (via \(B^{(l)}H^{(l)}\)), thus integrating both local and global information in the learning process.

\subsection{Multivariate-Temporal Convolutional Networks (MTCNs)}

Traditional Temporal Convolutional Neural Networks (TCNs) are a specialized variant of convolutional neural networks designed to process sequence data effectively. The core principle of TCNs is to apply a shared gated 1D convolution across a specified width \(w_l\) in the \(l^{th}\) layer, allowing the integration of information from \(w_l\) adjacent time points. In this study, we adapt traditional TCNs to simultaneously capture multivariate-temporal correlations. Specifically, we aim to incorporate information from all crime types at preceding time points for predicting a specific crime type at the current time point. This integration is facilitated by reshaping the multivariate and temporal dimensions together.

Each TCN layer \(H_l\) receives input from the preceding layer \(H_{l-1}\) and updates as follows:

\begin{equation}
H^{(l+1)} = f(\Gamma_l * H^{(l)} + b),
\end{equation}

where \(\Gamma_l\) represents the convolution filter for the layer, \(*\) denotes the shared convolution operation, and \(b\) represents the bias. If the previous hidden layer follows \(H_{l-1} \in \mathbb{R}^{B \times |V| \times (w_{l-1}C)}\), then the convolution filter is \(\Gamma_l \in \mathbb{R}^{(w_{l}C) \times (w_{l-1}C)}\), ensuring that \(H_l \in \mathbb{R}^{B \times |V| \times (w_{l}C)}\). Notably, if there is no padding in each TCN layer, it follows that \(w_l < w_{l-1}\).

The primary motivation for capturing multivariate-temporal correlations is to leverage the streamlined architecture of traditional TCNs for rapid training. Moreover, given that the temporal and multivariate dimensions typically exhibit shorter lengths compared to the spatial dimension, they can be treated as a single combined dimension. This approach helps mitigate the risk of extracting spurious relationships from the training data, which can lead to overfitting—a common issue when neural networks attempt to model relationships between two dimensions with extensive lengths~\cite{wang2023st}.

\subsection{General Structure}

\begin{figure}[h]
  \centering
  \includegraphics[width=\linewidth]{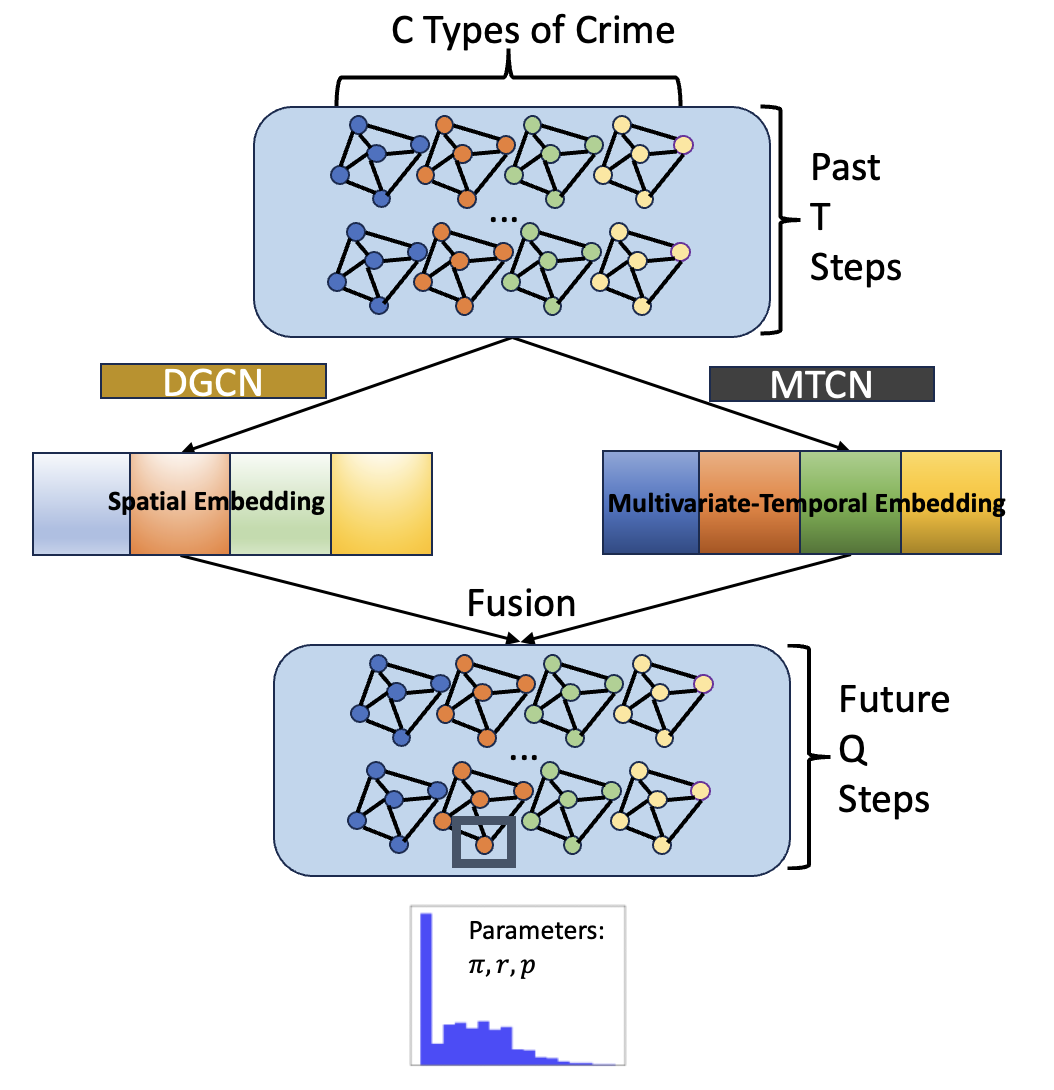}
  \caption{Framework of our STMGNN-ZINB method}
  \label{structure}
\end{figure}

As illustrated in Figure~\ref{structure}, we utilize DGCNs to capture spatial dependencies, resulting in spatial embeddings \(\pi_1\), \(p_1\), and \(r_1\). Concurrently, MTCNs are employed to capture multivariate temporal correlations, yielding multivariate temporal embeddings \(\pi_2\), \(p_2\), and \(r_2\). These embeddings are then fused into combined parameters \(\pi\), \(p\), and \(r\) using the Hadamard product. This method of integration is inspired by recent advances in uncertainty quantification techniques~\cite{zhuang2022uncertainty}.

To address the issue where zeros result in infinite values for the KL-divergence based variational lower bound~\cite{zhuang2022uncertainty}, we opt to directly utilize the negative likelihood as our loss function. This approach allows for a more accurate fitting of the distribution to the data, circumventing the limitations posed by zeros in the calculation of the KL-divergence.

\begin{equation}
 \text{NLL} = -\sum_{i=1}^{N} \begin{cases}
\log[\pi_i + (1 - \pi_i) \cdot (1 - p_i)^{r_i}] & \text{if } y_i = 0 \\
\log[(1 - \pi_i) \cdot \frac{\Gamma(r_i + y_i)}{\Gamma(r_i) \Gamma(y_i + 1)} p_i^{y_i} (1 - p_i)^{r_i}] & \text{if } y_i > 0
\end{cases} 
\end{equation}

\section{Experiments}
\label{exp}

\subsection{Experiment Setup}

\subsubsection{Data}

Following the methodology in~\cite{li2022spatial}, our experiments utilize two distinct crime datasets from New York City (NYC) and Chicago (CHI). These datasets cover various types of crime incidents, including Robbery and Larceny in NYC and Damage and Assault in CHI, across different locations within each city. For the purpose of our experiments, we divided each city into a spatial grid of 3 km × 3 km, resulting in 256 regions for NYC and 168 regions for CHI. Our prediction targets are set at a daily resolution.

The datasets were split into training and testing sets with a 7:1 ratio along the time dimension. Additionally, we used the last 30 days of the training set as a validation set to fine-tune the model parameters. Table~\ref{data} provides a summary of the dataset statistics.

\subsubsection{Evaluation Metrics}

We evaluate the performance of different models from three perspectives: 1. Point Estimation: We use Mean Absolute Error (MAE) to measure the accuracy of the mean value of the predicted distributions. A lower MAE indicates better accuracy. 2. Uncertainty Quantification: This includes Mean Prediction Interval Width (MPIW) and Prediction Interval Coverage Probability (PICP) within the 10\%-90\% confidence interval. MPIW calculates the average width of the confidence intervals, reflecting the extent of uncertainty in the predictions. PICP quantifies the percentage of actual data points that fall within these intervals, aiming for a coverage as close to 90\% as possible. Additionally, KL-Divergence is used to assess the similarity between the predicted and actual data distributions, with lower values indicating better model performance. 3. Discrete Metrics: We round our results to their closest integer to measure the true-zero rate and the F1-score. The true-zero rate evaluates the model’s ability to accurately represent data sparsity, whereas the F1-score assesses the accuracy of discrete predictions. Higher values of the true-zero rate and F1-score signify superior performance.

\subsubsection{Baseline Methods}

In order to explore the advantages of STMGNN-ZINB, we compare the STMGNN-ZINB results against three other models: (1) Historical Value (HA) serves as the statistical baseline. It uses historical input directly to represent prediction results. (2) Spatial-Temporal Graph Convolutional Networks (STGCN)~\footnote{https://github.com/FelixOpolka/STGCN-PyTorch} is the state-of-the-art deep learning model for spatial-temporal prediction but it only produces point estimates; (3) Models with probabilistic assumptions to replace ZINB: Negative Binomial (STMGNN-NB), Gaussian (STMGNN-G), and Truncated Normal (STMGNN-TN).

\begin{table}
    \centering
    \resizebox{\columnwidth}{!}{
    {\small 
    \begin{tabular}{|c||c|c|c|c|c|} \hline
        Dataset & Time Span & Category & Count & \multicolumn{2}{c|}{Zero Rate} \\ \hline
        \multirow{3}{*}{NYC-Crimes} & \multirow{3}{*}{Jan. 2014 to Dec. 2015} & Burglary & 31799 & \multicolumn{2}{c|}{89} \\ \cline{3-6}
        & & Larceny & 85899 & \multicolumn{2}{c|}{81} \\ \cline{3-6}
        & & Robbery & 33453 & \multicolumn{2}{c|}{89} \\ \cline{3-6}
        & & Assault & 40429 & \multicolumn{2}{c|}{88} \\ \hline
        \multirow{4}{*}{CHI-Crimes} & \multirow{3}{*}{Jan. 2016 to Dec. 2017} & Theft & 124630 & \multicolumn{2}{c|}{68} \\ \cline{3-6}
        & & Battery & 99389 & \multicolumn{2}{c|}{71} \\ \cline{3-6}
        & & Assault & 37972 & \multicolumn{2}{c|}{81} \\ \cline{3-6}
        & & Damage & 59886 & \multicolumn{2}{c|}{75} \\ \hline
    \end{tabular}
    }
    }
    \caption{Statistics of Experimented Urban Crime Datasets}
    \label{data}
\end{table}

\begin{table*}[!ht]
    \centering
    \renewcommand{\arraystretch}{0.95} 
    {\fontsize{9}{11}\selectfont
    \begin{tabular}{|c|c|c|c|c|c|c|c|}
        \hline
        Datasets & Methods & MAE & KL Divergence & PICP & MPIW & F1 Score & True Zero Rate \\
        \hline
        \hline
            \multicolumn{2}{|c|}{} & \multicolumn{1}{c|}{Point Estimation} & \multicolumn{3}{c|}{Uncertainty Metrics} & \multicolumn{2}{c|}{Discrete Metrics} \\
        
        \hline
        \hline

                \multicolumn{1}{|c|}{NYC-Crimes} & \multicolumn{1}{c|}{STMGNN-ZINB} 
        & \multicolumn{1}{c|}{\textbf{0.2128}}
        & \multicolumn{1}{c|}{\underline{0.6147}} & \multicolumn{1}{c|}{\textbf{0.9505\%}} 
        & \multicolumn{1}{c|}{\textbf{1.028}}
        & \multicolumn{1}{c|}{\underline{0.6556}} & \multicolumn{1}{c|}{\underline{96.57\%}} \\

        \multicolumn{1}{|c|}{} & \multicolumn{1}{c|}{STMGNN-G} 
        & \multicolumn{1}{c|}{0.2536} 
        & \multicolumn{1}{c|}{\textbf{0.1148}} & \multicolumn{1}{c|}{\underline{0.9557\%}}
        & \multicolumn{1}{c|}{\underline{1.30}}
        & \multicolumn{1}{c|}{<0.1} & \multicolumn{1}{c|}{\textbf{100.0\%}} \\

        \multicolumn{1}{|c|}{} & \multicolumn{1}{c|}{STNGNN-NB} 
        & \multicolumn{1}{c|}{\underline{0.2300}}
        & \multicolumn{1}{c|}{1.0218} & \multicolumn{1}{c|}{0.9690\%} 
        & \multicolumn{1}{c|}{\underline{1.30}}
        & \multicolumn{1}{c|}{\textbf{0.6644}} & \multicolumn{1}{c|}{93.01\%} \\

        \multicolumn{1}{|c|}{} & \multicolumn{1}{c|}{STMGNN-TN} 
        & \multicolumn{1}{c|}{0.3206} 
        & \multicolumn{1}{c|}{1.0636} & \multicolumn{1}{c|}{0.9682\%} 
        & \multicolumn{1}{c|}{1.5148}
        & \multicolumn{1}{c|}{0.5638} & \multicolumn{1}{c|}{90.92\%} \\

        \multicolumn{1}{|c|}{} & \multicolumn{1}{c|}{HV} 
        & \multicolumn{1}{c|}{0.3014} 
        & \multicolumn{1}{c|}{1.2783} & \multicolumn{1}{c|}{/} 
        & \multicolumn{1}{c|}{/}
        & \multicolumn{1}{c|}{0.5234} & \multicolumn{1}{c|}{92.53\%}\\%

         \multicolumn{1}{|c|}{} & \multicolumn{1}{c|}{STGCN} 
        & \multicolumn{1}{c|}{0.2527} 
        & \multicolumn{1}{c|}{2.0259} & \multicolumn{1}{c|}{/} 
        & \multicolumn{1}{c|}{/}
        & \multicolumn{1}{c|}{0.6537} & \multicolumn{1}{c|}{87.10\%}\\

        \hline

        \multicolumn{1}{|c|}{CHI-Crimes} & \multicolumn{1}{c|}{STMGNN-ZINB} 
      & \multicolumn{1}{c|}{\textbf{0.5242}} 
        & \multicolumn{1}{c|}{\textbf{1.3945}} & \multicolumn{1}{c|}{\textbf{0.9141\%}} 
        & \multicolumn{1}{c|}{\textbf{2.134}}
        & \multicolumn{1}{c|}{0.7693} & \multicolumn{1}{c|}{\textbf{90.73\%}} \\

        \multicolumn{1}{|c|}{} & \multicolumn{1}{c|}{STMGNN-G} 
        & \multicolumn{1}{c|}{0.5587} 
        & \multicolumn{1}{c|}{\underline{1.8047}} & \multicolumn{1}{c|}{\underline{0.8938\%} }
        & \multicolumn{1}{c|}{2.532}
        & \multicolumn{1}{c|}{0.7259} & \multicolumn{1}{c|}{86.50\%} \\

        \multicolumn{1}{|c|}{} & \multicolumn{1}{c|}{STMGNN-NB} 
        & \multicolumn{1}{c|}{\underline{0.5575} }
        & \multicolumn{1}{c|}{2.0259} & \multicolumn{1}{c|}{0.9444\%} 
        & \multicolumn{1}{c|}{\underline{2.4164}}
        & \multicolumn{1}{c|}{\underline{0.7769}} & \multicolumn{1}{c|}{87.10\%}\\

        \multicolumn{1}{|c|}{} & \multicolumn{1}{c|}{STMGNN-TN} 
        & \multicolumn{1}{c|}{0.7619} 
        & \multicolumn{1}{c|}{2.7606} & \multicolumn{1}{c|}{0.8617\%} 
        & \multicolumn{1}{c|}{2.906}
        & \multicolumn{1}{c|}{0.6329} & \multicolumn{1}{c|}{61.13\%} \\

        \multicolumn{1}{|c|}{} & \multicolumn{1}{c|}{HV} 
        & \multicolumn{1}{c|}{0.6560} 
        & \multicolumn{1}{c|}{2.2312} & \multicolumn{1}{c|}{/} 
        & \multicolumn{1}{c|}{/}
        & \multicolumn{1}{c|}{0.6849} & \multicolumn{1}{c|}{\underline{89.09\%}}\\%
         
         \multicolumn{1}{|c|}{} & \multicolumn{1}{c|}{STGCN} 
        & \multicolumn{1}{c|}{0.5535} 
        & \multicolumn{1}{c|}{1.9512} & \multicolumn{1}{c|}{/} 
        & \multicolumn{1}{c|}{/}
        & \multicolumn{1}{c|}{\textbf{0.7858}} & \multicolumn{1}{c|}{86.59\%}\\
       \hline
    \end{tabular}
    }
    \caption{Experimental Results in NYC-Crimes and CHI-Crimes. The best and second-best scores are bold and underlined.}
\label{result}
   
\end{table*}

\subsection{Experimental Results and Analysis}

As shown in Table~\ref{result}, STMGNN-ZINB outperforms all baseline methods across the majority of evaluation metrics. Specifically, in point estimation, STMGNN-ZINB achieves the lowest Mean Absolute Error (MAE), indicating that its mean value accurately reflects the trends in real crime data. For uncertainty metrics, the lower Mean Prediction Interval Width (MPIW) of STMGNN-ZINB suggests that our model generates predictions with less uncertainty and greater precision. STMGNN-ZINB surpasses other methods on metrics such as the true zero rate and F1 Score in most cases, signifying STMGNN-ZINB’s efficiency in capturing zero values.

During our experiments, we observed that STMGNN-G performed exceptionally well in terms of KL Divergence and True Zero Rate in the New York City dataset. However, its F1 Score was notably low. Further investigation revealed that STMGNN-G predominantly outputs near-zero mean values in this dataset. Given that the New York City dataset has a significantly higher zero rate compared to the Chicago dataset, the low F1 Score suggests that STMGNN-G struggles to effectively predict crime occurrences. Hence, we conclude that it is crucial to evaluate a model using diverse and multiple evaluation metrics. The performance of STMGNN-ZINB is more comprehensive than that of STMGNN-G.

\begin{figure}[h]
  \centering
  \includegraphics[width=0.5\linewidth]{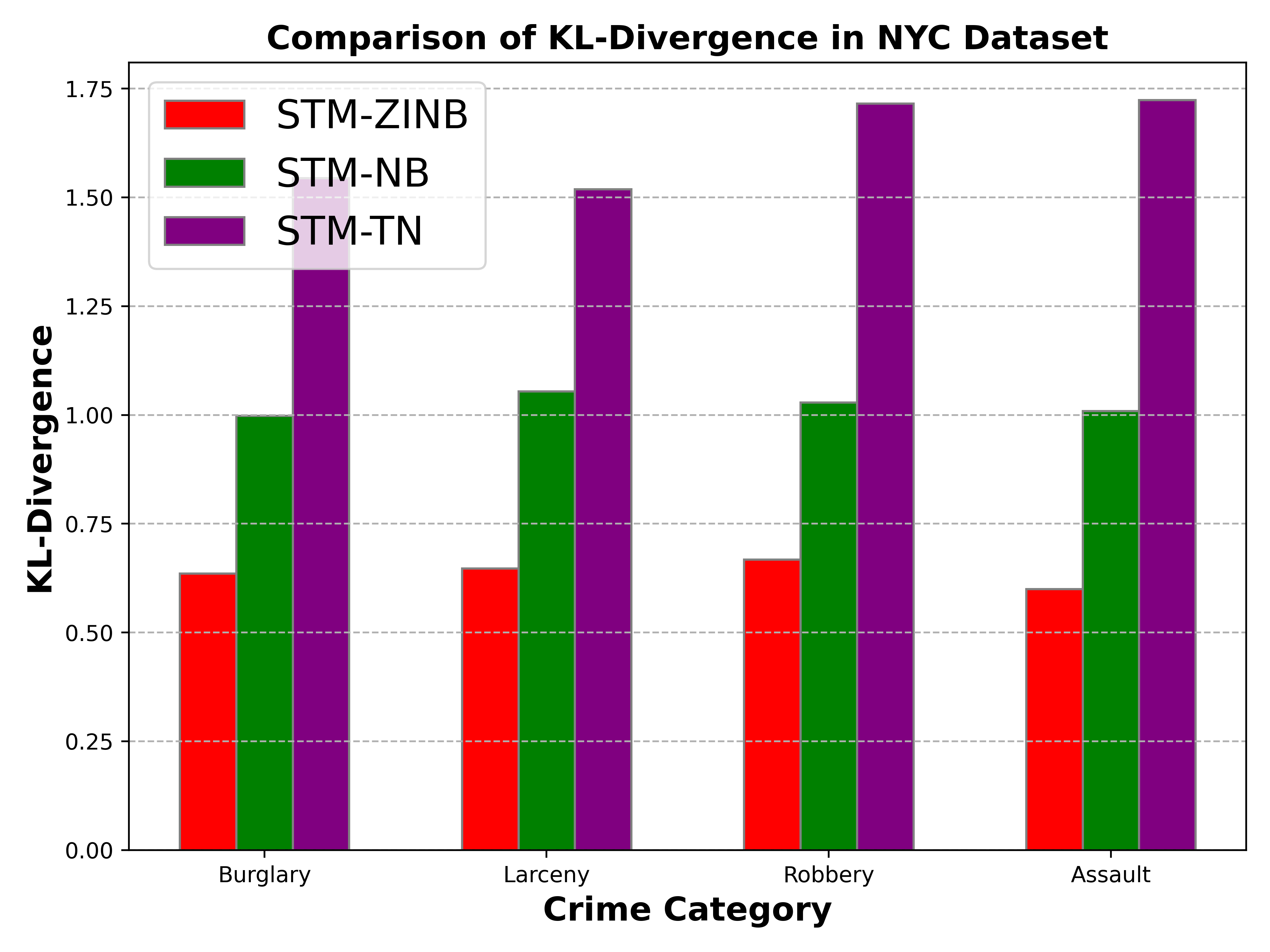}
  \caption{Comparison of KL-Divergence in NYC Dataset}
  \label{kl1}
\end{figure}

\begin{figure}[h]
  \centering
  \includegraphics[width=0.5\linewidth]{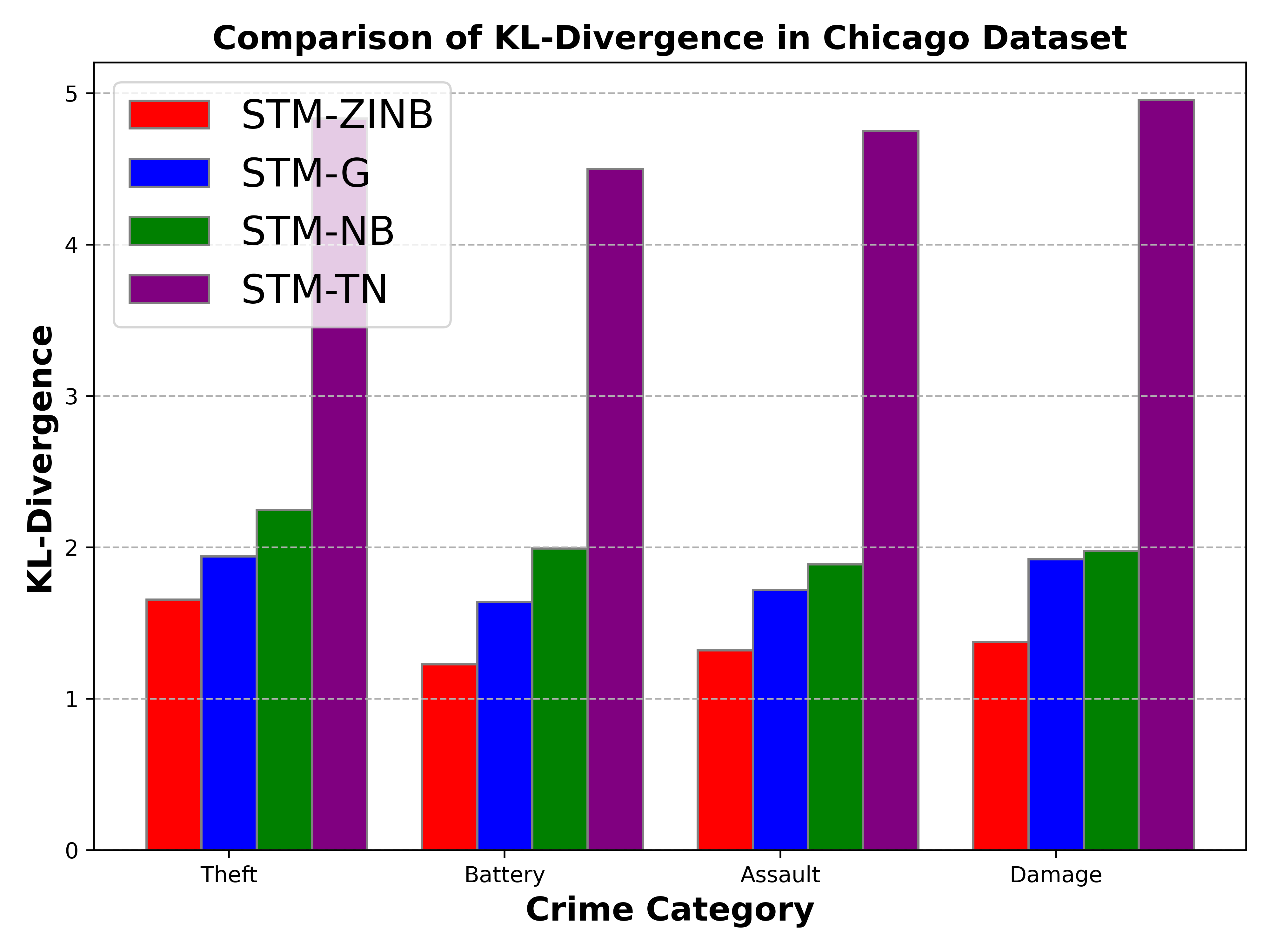}
  \caption{Comparison of KL-Divergence in Chicago Dataset}
  \label{kl2}
\end{figure}

When excluding STMGNN-G from the New York City dataset to evaluate KL divergence, as illustrated in Figures~\ref{kl1} and~\ref{kl2}, we can conclude that STMGNN-ZINB demonstrates the best performance in approximating the actual distribution in both datasets.

\subsection{Interpretation of Parameter \(\pi\)}

The sparsity parameter \(\pi\) in the Zero-Inflated Negative Binomial (ZINB) distribution quantifies the likelihood that a specific zone is devoid of any crime activities. Our STMGNN-ZINB model outputs \(\pi\) for various types of crime activities across different spatial regions for successive future time steps.

\begin{figure}[h]
\begin{minipage}{0.25\textwidth}
    \includegraphics[width=\linewidth]{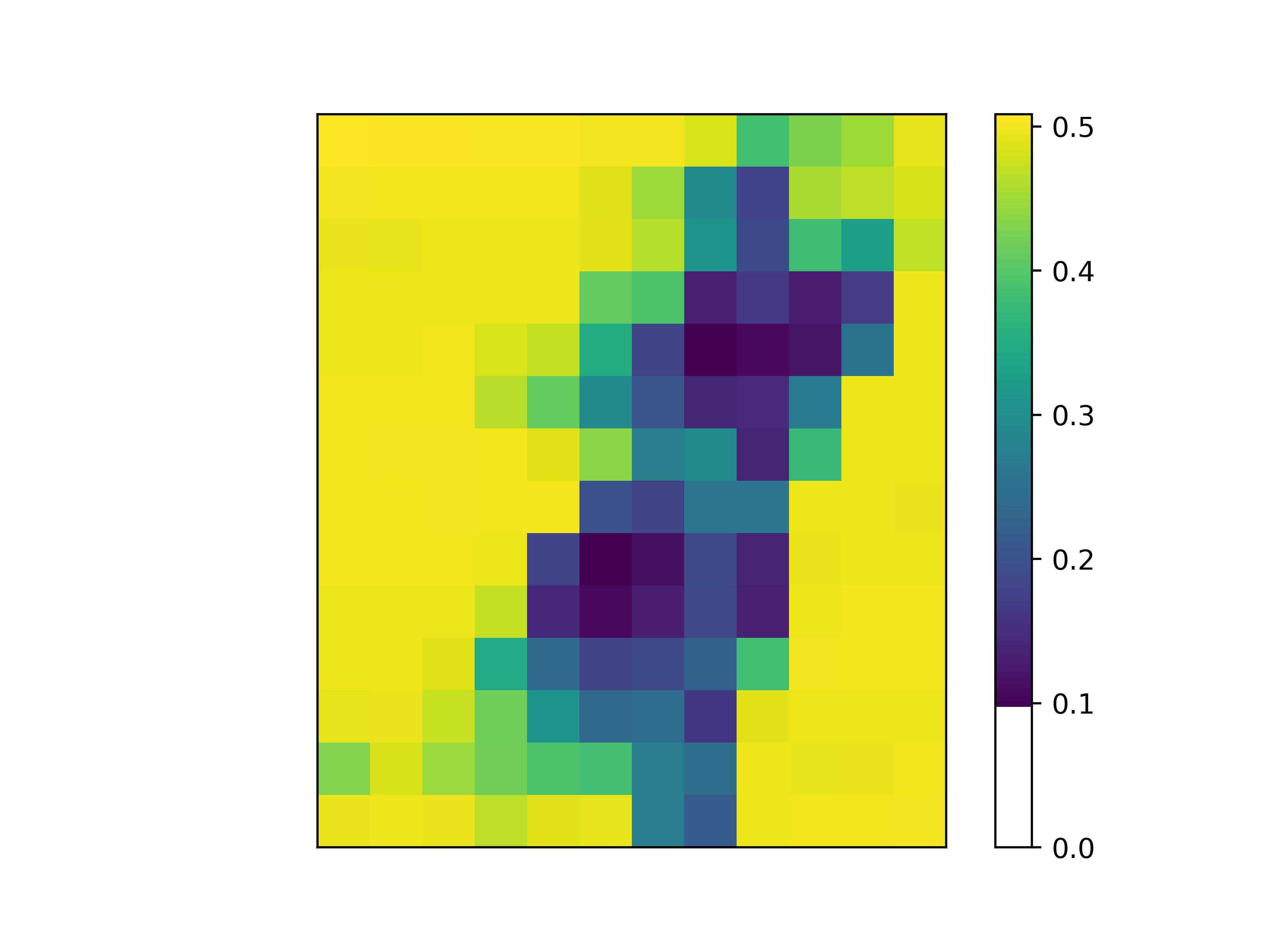} 
    \label{fig:image1}
\end{minipage}%
\begin{minipage}{0.25\textwidth}
    \includegraphics[width=\linewidth]{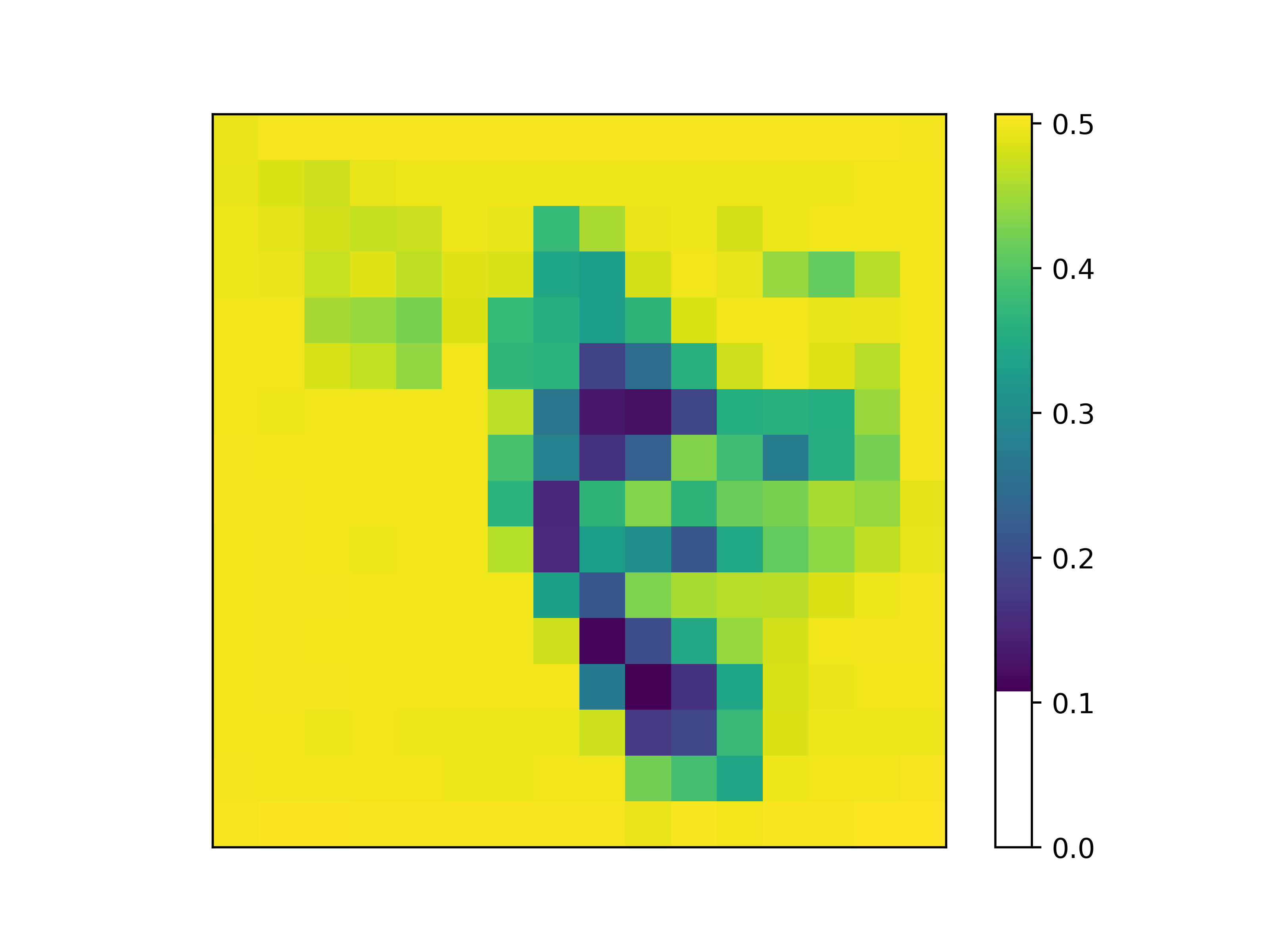}
\end{minipage}
\caption{Heatmap of \(\pi\) in the entire urban space of NYC (left) and Chicago (right). Results are averaged over the multivariate dimension and temporal dimension.}
\label{fig:image2}
\end{figure}

Figure~\ref{fig:image2} displays a heatmap of the overall crime activities (parameter \(\pi\)) for New York City and Chicago. This visualization offers valuable insights for government agencies and public security entities. Notably, it reveals the presence of spatial locality, where changes in crime activities gradually vary across the spatial dimension. It is uncommon to find isolated small regions with low crime rates surrounded by areas with high crime rates. Furthermore, consistent with mainstream criminological research, our findings indicate a higher concentration of crime activities in the downtown areas of both cities compared to the marginal regions (rural areas)~\cite{lan2021spatial,rosenfeld2020pandemic}. This pattern underscores the heightened crime rates typically observed in urban centers as opposed to rural settings. Thus, the sparsity parameter \(\pi\) significantly enhances the interpretability of the STMGNN-ZINB model, offering a more nuanced understanding compared to models based on simpler distribution assumptions.

\section{Conclusion}
\label{con}

In this work, we introduce a novel combination of Spatial Temporal Multivariate Graph Neural Networks and Zero-Inflated Negative Binomial distribution to quantify uncertainty in the urban crime prediction problem. We validated our model's performance through extensive experiments across five representative scenarios, focusing particularly on point estimation and uncertainty measurement. Given that the parameter \(\pi\) has a clear physical interpretation, our model could assist social security decision-makers in efficiently allocating limited social resources to different regions. This approach not only enhances predictive accuracy but also aids in strategic planning and resource management.

\bibliographystyle{ACM-Reference-Format}
\bibliography{sample-base}


\begin{thebibliography}{29}


\ifx \showCODEN    \undefined \def \showCODEN     #1{\unskip}     \fi
\ifx \showDOI      \undefined \def \showDOI       #1{#1}\fi
\ifx \showISBNx    \undefined \def \showISBNx     #1{\unskip}     \fi
\ifx \showISBNxiii \undefined \def \showISBNxiii  #1{\unskip}     \fi
\ifx \showISSN     \undefined \def \showISSN      #1{\unskip}     \fi
\ifx \showLCCN     \undefined \def \showLCCN      #1{\unskip}     \fi
\ifx \shownote     \undefined \def \shownote      #1{#1}          \fi
\ifx \showarticletitle \undefined \def \showarticletitle #1{#1}   \fi
\ifx \showURL      \undefined \def \showURL       {\relax}        \fi
\providecommand\bibfield[2]{#2}
\providecommand\bibinfo[2]{#2}
\providecommand\natexlab[1]{#1}
\providecommand\showeprint[2][]{arXiv:#2}

\bibitem[Almanie et~al\mbox{.}(2015)]%
        {almanie2015crime}
\bibfield{author}{\bibinfo{person}{Tahani Almanie}, \bibinfo{person}{Rsha Mirza}, {and} \bibinfo{person}{Elizabeth Lor}.} \bibinfo{year}{2015}\natexlab{}.
\newblock \showarticletitle{Crime prediction based on crime types and using spatial and temporal criminal hotspots}.
\newblock \bibinfo{journal}{\emph{arXiv preprint arXiv:1508.02050}} (\bibinfo{year}{2015}).
\newblock


\bibitem[Bautista-Montesano et~al\mbox{.}(2022)]%
        {bautista2022autonomous}
\bibfield{author}{\bibinfo{person}{Rolando Bautista-Montesano}, \bibinfo{person}{Renato Galluzzi}, \bibinfo{person}{Kangrui Ruan}, \bibinfo{person}{Yongjie Fu}, {and} \bibinfo{person}{Xuan Di}.} \bibinfo{year}{2022}\natexlab{}.
\newblock \showarticletitle{Autonomous navigation at unsignalized intersections: A coupled reinforcement learning and model predictive control approach}.
\newblock \bibinfo{journal}{\emph{Transportation research part C: emerging technologies}}  \bibinfo{volume}{139} (\bibinfo{year}{2022}), \bibinfo{pages}{103662}.
\newblock


\bibitem[Ding et~al\mbox{.}(2017)]%
        {ding2017using}
\bibfield{author}{\bibinfo{person}{Chuan Ding}, \bibinfo{person}{Jinxiao Duan}, \bibinfo{person}{Yanru Zhang}, \bibinfo{person}{Xinkai Wu}, {and} \bibinfo{person}{Guizhen Yu}.} \bibinfo{year}{2017}\natexlab{}.
\newblock \showarticletitle{Using an ARIMA-GARCH modeling approach to improve subway short-term ridership forecasting accounting for dynamic volatility}.
\newblock \bibinfo{journal}{\emph{IEEE Transactions on Intelligent Transportation Systems}} \bibinfo{volume}{19}, \bibinfo{number}{4} (\bibinfo{year}{2017}), \bibinfo{pages}{1054--1064}.
\newblock


\bibitem[Gao et~al\mbox{.}(2023)]%
        {gao2023spatiotemporal}
\bibfield{author}{\bibinfo{person}{Xiaowei Gao}, \bibinfo{person}{Xinke Jiang}, \bibinfo{person}{Dingyi Zhuang}, \bibinfo{person}{Huanfa Chen}, \bibinfo{person}{Shenhao Wang}, {and} \bibinfo{person}{James Haworth}.} \bibinfo{year}{2023}\natexlab{}.
\newblock \showarticletitle{Spatiotemporal graph neural networks with uncertainty quantification for traffic incident risk prediction}.
\newblock \bibinfo{journal}{\emph{arXiv preprint arXiv:2309.05072}} (\bibinfo{year}{2023}).
\newblock


\bibitem[Guo et~al\mbox{.}(2014)]%
        {guo2014adaptive}
\bibfield{author}{\bibinfo{person}{Jianhua Guo}, \bibinfo{person}{Wei Huang}, {and} \bibinfo{person}{Billy~M Williams}.} \bibinfo{year}{2014}\natexlab{}.
\newblock \showarticletitle{Adaptive Kalman filter approach for stochastic short-term traffic flow rate prediction and uncertainty quantification}.
\newblock \bibinfo{journal}{\emph{Transportation Research Part C: Emerging Technologies}}  \bibinfo{volume}{43} (\bibinfo{year}{2014}), \bibinfo{pages}{50--64}.
\newblock


\bibitem[He and Jiang(2023a)]%
        {he2023survey}
\bibfield{author}{\bibinfo{person}{Wenchong He} {and} \bibinfo{person}{Zhe Jiang}.} \bibinfo{year}{2023}\natexlab{a}.
\newblock \showarticletitle{A survey on uncertainty quantification methods for deep neural networks: An uncertainty source perspective}.
\newblock \bibinfo{journal}{\emph{arXiv preprint arXiv:2302.13425}} (\bibinfo{year}{2023}).
\newblock


\bibitem[He and Jiang(2023b)]%
        {he2023uncertainty}
\bibfield{author}{\bibinfo{person}{Wenchong He} {and} \bibinfo{person}{Zhe Jiang}.} \bibinfo{year}{2023}\natexlab{b}.
\newblock \showarticletitle{Uncertainty Quantification of Deep Learning for Spatiotemporal Data: Challenges and Opportunities}.
\newblock \bibinfo{journal}{\emph{arXiv preprint arXiv:2311.02485}} (\bibinfo{year}{2023}).
\newblock


\bibitem[Huang et~al\mbox{.}(2019)]%
        {huang2019mist}
\bibfield{author}{\bibinfo{person}{Chao Huang}, \bibinfo{person}{Chuxu Zhang}, \bibinfo{person}{Jiashu Zhao}, \bibinfo{person}{Xian Wu}, \bibinfo{person}{Dawei Yin}, {and} \bibinfo{person}{Nitesh Chawla}.} \bibinfo{year}{2019}\natexlab{}.
\newblock \showarticletitle{Mist: A multiview and multimodal spatial-temporal learning framework for citywide abnormal event forecasting}. In \bibinfo{booktitle}{\emph{The world wide web conference}}. \bibinfo{pages}{717--728}.
\newblock


\bibitem[Huang et~al\mbox{.}(2018)]%
        {huang2018deepcrime}
\bibfield{author}{\bibinfo{person}{Chao Huang}, \bibinfo{person}{Junbo Zhang}, \bibinfo{person}{Yu Zheng}, {and} \bibinfo{person}{Nitesh~V Chawla}.} \bibinfo{year}{2018}\natexlab{}.
\newblock \showarticletitle{DeepCrime: Attentive hierarchical recurrent networks for crime prediction}. In \bibinfo{booktitle}{\emph{Proceedings of the 27th ACM international conference on information and knowledge management}}. \bibinfo{pages}{1423--1432}.
\newblock


\bibitem[Jiang and Zhang(2018)]%
        {jiang2018sparse}
\bibfield{author}{\bibinfo{person}{Mengmeng Jiang} {and} \bibinfo{person}{Hang Zhang}.} \bibinfo{year}{2018}\natexlab{}.
\newblock \showarticletitle{Sparse estimation in high-dimensional zero-inflated Poisson regression model}. In \bibinfo{booktitle}{\emph{Journal of Physics: Conference Series}}, Vol.~\bibinfo{volume}{1053}. IOP Publishing, \bibinfo{pages}{012128}.
\newblock


\bibitem[Jiang et~al\mbox{.}(2023)]%
        {jiang2023uncertainty}
\bibfield{author}{\bibinfo{person}{Xinke Jiang}, \bibinfo{person}{Dingyi Zhuang}, \bibinfo{person}{Xianghui Zhang}, \bibinfo{person}{Hao Chen}, \bibinfo{person}{Jiayuan Luo}, {and} \bibinfo{person}{Xiaowei Gao}.} \bibinfo{year}{2023}\natexlab{}.
\newblock \showarticletitle{Uncertainty quantification via spatial-temporal tweedie model for zero-inflated and long-tail travel demand prediction}. In \bibinfo{booktitle}{\emph{Proceedings of the 32nd ACM International Conference on Information and Knowledge Management}}. \bibinfo{pages}{3983--3987}.
\newblock


\bibitem[Ke et~al\mbox{.}(2020)]%
        {ke2020enhancing}
\bibfield{author}{\bibinfo{person}{Zemian Ke}, \bibinfo{person}{Zhibin Li}, \bibinfo{person}{Zehong Cao}, {and} \bibinfo{person}{Pan Liu}.} \bibinfo{year}{2020}\natexlab{}.
\newblock \showarticletitle{Enhancing transferability of deep reinforcement learning-based variable speed limit control using transfer learning}.
\newblock \bibinfo{journal}{\emph{IEEE Transactions on Intelligent Transportation Systems}} \bibinfo{volume}{22}, \bibinfo{number}{7} (\bibinfo{year}{2020}), \bibinfo{pages}{4684--4695}.
\newblock


\bibitem[Lan et~al\mbox{.}(2021)]%
        {lan2021spatial}
\bibfield{author}{\bibinfo{person}{Minxuan Lan}, \bibinfo{person}{Lin Liu}, {and} \bibinfo{person}{John~E Eck}.} \bibinfo{year}{2021}\natexlab{}.
\newblock \showarticletitle{A spatial analytical approach to assess the impact of a casino on crime: An example of JACK Casino in downtown Cincinnati}.
\newblock \bibinfo{journal}{\emph{Cities}}  \bibinfo{volume}{111} (\bibinfo{year}{2021}), \bibinfo{pages}{103003}.
\newblock


\bibitem[Li et~al\mbox{.}(2022)]%
        {li2022spatial}
\bibfield{author}{\bibinfo{person}{Zhonghang Li}, \bibinfo{person}{Chao Huang}, \bibinfo{person}{Lianghao Xia}, \bibinfo{person}{Yong Xu}, {and} \bibinfo{person}{Jian Pei}.} \bibinfo{year}{2022}\natexlab{}.
\newblock \showarticletitle{Spatial-temporal hypergraph self-supervised learning for crime prediction}. In \bibinfo{booktitle}{\emph{2022 IEEE 38th International Conference on Data Engineering (ICDE)}}. IEEE, \bibinfo{pages}{2984--2996}.
\newblock


\bibitem[Lin et~al\mbox{.}(2023)]%
        {lin2023mmst}
\bibfield{author}{\bibinfo{person}{Fudong Lin}, \bibinfo{person}{Summer Crawford}, \bibinfo{person}{Kaleb Guillot}, \bibinfo{person}{Yihe Zhang}, \bibinfo{person}{Yan Chen}, \bibinfo{person}{Xu Yuan}, \bibinfo{person}{Li Chen}, \bibinfo{person}{Shelby Williams}, \bibinfo{person}{Robert Minvielle}, \bibinfo{person}{Xiangming Xiao}, {et~al\mbox{.}}} \bibinfo{year}{2023}\natexlab{}.
\newblock \showarticletitle{MMST-ViT: Climate Change-aware Crop Yield Prediction via Multi-Modal Spatial-Temporal Vision Transformer}. In \bibinfo{booktitle}{\emph{Proceedings of the IEEE/CVF International Conference on Computer Vision}}. \bibinfo{pages}{5774--5784}.
\newblock


\bibitem[Liu et~al\mbox{.}(2024)]%
        {liu2024enhancing}
\bibfield{author}{\bibinfo{person}{Yinan Liu}, \bibinfo{person}{Xinyu Dong}, \bibinfo{person}{Weimin Lyu}, \bibinfo{person}{Richard~N Rosenthal}, \bibinfo{person}{Rachel Wong}, \bibinfo{person}{Tengfei Ma}, \bibinfo{person}{Jun Kong}, {and} \bibinfo{person}{Fusheng Wang}.} \bibinfo{year}{2024}\natexlab{}.
\newblock \showarticletitle{Enhancing Clinical Predictive Modeling through Model Complexity-Driven Class Proportion Tuning for Class Imbalanced Data: An Empirical Study on Opioid Overdose Prediction}.
\newblock \bibinfo{journal}{\emph{AMIA Summits on Translational Science Proceedings}}  \bibinfo{volume}{2024} (\bibinfo{year}{2024}), \bibinfo{pages}{334}.
\newblock


\bibitem[Mo et~al\mbox{.}(2024)]%
        {mo2024cross}
\bibfield{author}{\bibinfo{person}{Zhaobin Mo}, \bibinfo{person}{Haotian Xiang}, {and} \bibinfo{person}{Xuan Di}.} \bibinfo{year}{2024}\natexlab{}.
\newblock \showarticletitle{Cross-and Context-Aware Attention Based Spatial-Temporal Graph Convolutional Networks for Human Mobility Prediction}.
\newblock \bibinfo{journal}{\emph{ACM Transactions on Spatial Algorithms and Systems}} (\bibinfo{year}{2024}).
\newblock


\bibitem[Rosenfeld and Lopez(2020)]%
        {rosenfeld2020pandemic}
\bibfield{author}{\bibinfo{person}{Richard Rosenfeld} {and} \bibinfo{person}{Ernesto Lopez}.} \bibinfo{year}{2020}\natexlab{}.
\newblock \showarticletitle{Pandemic, social unrest, and crime in US cities}.
\newblock \bibinfo{journal}{\emph{Federal Sentencing Reporter}} \bibinfo{volume}{33}, \bibinfo{number}{1/2} (\bibinfo{year}{2020}), \bibinfo{pages}{72--82}.
\newblock


\bibitem[Ruan and Di(2024)]%
        {ruan2024infostgcan}
\bibfield{author}{\bibinfo{person}{Kangrui Ruan} {and} \bibinfo{person}{Xuan Di}.} \bibinfo{year}{2024}\natexlab{}.
\newblock \showarticletitle{InfoSTGCAN: An Information-Maximizing Spatial-Temporal Graph Convolutional Attention Network for Heterogeneous Human Trajectory Prediction}.
\newblock \bibinfo{journal}{\emph{Computers}} \bibinfo{volume}{13}, \bibinfo{number}{6} (\bibinfo{year}{2024}), \bibinfo{pages}{151}.
\newblock


\bibitem[Safat et~al\mbox{.}(2021)]%
        {safat2021empirical}
\bibfield{author}{\bibinfo{person}{Wajiha Safat}, \bibinfo{person}{Sohail Asghar}, {and} \bibinfo{person}{Saira~Andleeb Gillani}.} \bibinfo{year}{2021}\natexlab{}.
\newblock \showarticletitle{Empirical analysis for crime prediction and forecasting using machine learning and deep learning techniques}.
\newblock \bibinfo{journal}{\emph{IEEE access}}  \bibinfo{volume}{9} (\bibinfo{year}{2021}), \bibinfo{pages}{70080--70094}.
\newblock


\bibitem[Sun et~al\mbox{.}(2021)]%
        {sun2021crimeforecaster}
\bibfield{author}{\bibinfo{person}{Jiao Sun}, \bibinfo{person}{Mingxuan Yue}, \bibinfo{person}{Zongyu Lin}, \bibinfo{person}{Xiaochen Yang}, \bibinfo{person}{Luciano Nocera}, \bibinfo{person}{Gabriel Kahn}, {and} \bibinfo{person}{Cyrus Shahabi}.} \bibinfo{year}{2021}\natexlab{}.
\newblock \showarticletitle{Crimeforecaster: Crime prediction by exploiting the geographical neighborhoods’ spatiotemporal dependencies}. In \bibinfo{booktitle}{\emph{Machine Learning and Knowledge Discovery in Databases. Applied Data Science and Demo Track: European Conference, ECML PKDD 2020, Ghent, Belgium, September 14--18, 2020, Proceedings, Part V}}. Springer, \bibinfo{pages}{52--67}.
\newblock


\bibitem[Wang et~al\mbox{.}(2022a)]%
        {wang2022hagen}
\bibfield{author}{\bibinfo{person}{Chenyu Wang}, \bibinfo{person}{Zongyu Lin}, \bibinfo{person}{Xiaochen Yang}, \bibinfo{person}{Jiao Sun}, \bibinfo{person}{Mingxuan Yue}, {and} \bibinfo{person}{Cyrus Shahabi}.} \bibinfo{year}{2022}\natexlab{a}.
\newblock \showarticletitle{Hagen: Homophily-aware graph convolutional recurrent network for crime forecasting}. In \bibinfo{booktitle}{\emph{Proceedings of the AAAI Conference on Artificial Intelligence}}, Vol.~\bibinfo{volume}{36}. \bibinfo{pages}{4193--4200}.
\newblock


\bibitem[Wang et~al\mbox{.}(2022b)]%
        {wang2022novel}
\bibfield{author}{\bibinfo{person}{Zepu Wang}, \bibinfo{person}{Peng Sun}, {and} \bibinfo{person}{Azzedine Boukerche}.} \bibinfo{year}{2022}\natexlab{b}.
\newblock \showarticletitle{A Novel Time Efficient Machine Learning-based Traffic Flow Prediction Method for Large Scale Road Network}. In \bibinfo{booktitle}{\emph{Proceedings of the 2022 IEEE International Conference on Communications}}. IEEE, \bibinfo{pages}{3532--3537}.
\newblock


\bibitem[Wang et~al\mbox{.}(2022c)]%
        {wang2022novel1}
\bibfield{author}{\bibinfo{person}{Zepu Wang}, \bibinfo{person}{Peng Sun}, \bibinfo{person}{Yulin Hu}, {and} \bibinfo{person}{Azzedine Boukerche}.} \bibinfo{year}{2022}\natexlab{c}.
\newblock \showarticletitle{A Novel Mixed Method of Machine Learning Based Models in Vehicular Traffic Flow Prediction}. In \bibinfo{booktitle}{\emph{Proceedings of the 25th International ACM Conference on Modeling Analysis And Simulation of Wireless And Mobile Systems}}. \bibinfo{pages}{95--101}.
\newblock


\bibitem[Wang et~al\mbox{.}(2023)]%
        {wang2023st}
\bibfield{author}{\bibinfo{person}{Zepu Wang}, \bibinfo{person}{Dingyi Zhuang}, \bibinfo{person}{Yankai Li}, \bibinfo{person}{Jinhua Zhao}, {and} \bibinfo{person}{Peng Sun}.} \bibinfo{year}{2023}\natexlab{}.
\newblock \showarticletitle{ST-GIN: An Uncertainty Quantification Approach in Traffic Data Imputation with Spatio-temporal Graph Attention And Bidirectional Recurrent United Neural Networks}.
\newblock \bibinfo{journal}{\emph{arXiv preprint: 2305.06480}} (\bibinfo{year}{2023}).
\newblock


\bibitem[Wu et~al\mbox{.}(2020)]%
        {wu2020hierarchically}
\bibfield{author}{\bibinfo{person}{Xian Wu}, \bibinfo{person}{Chao Huang}, \bibinfo{person}{Chuxu Zhang}, {and} \bibinfo{person}{Nitesh~V Chawla}.} \bibinfo{year}{2020}\natexlab{}.
\newblock \showarticletitle{Hierarchically structured transformer networks for fine-grained spatial event forecasting}. In \bibinfo{booktitle}{\emph{Proceedings of the web conference 2020}}. \bibinfo{pages}{2320--2330}.
\newblock


\bibitem[Zhao et~al\mbox{.}(2016)]%
        {zhao2016multi}
\bibfield{author}{\bibinfo{person}{Liang Zhao}, \bibinfo{person}{Feng Chen}, \bibinfo{person}{Chang-Tien Lu}, {and} \bibinfo{person}{Naren Ramakrishnan}.} \bibinfo{year}{2016}\natexlab{}.
\newblock \showarticletitle{Multi-resolution spatial event forecasting in social media}. In \bibinfo{booktitle}{\emph{2016 IEEE 16th International Conference on Data Mining (ICDM)}}. IEEE, \bibinfo{pages}{689--698}.
\newblock


\bibitem[Zhuang et~al\mbox{.}(2023)]%
        {zhuang2023sauc}
\bibfield{author}{\bibinfo{person}{Dingyi Zhuang}, \bibinfo{person}{Yuheng Bu}, \bibinfo{person}{Guang Wang}, \bibinfo{person}{Shenhao Wang}, {and} \bibinfo{person}{Jinhua Zhao}.} \bibinfo{year}{2023}\natexlab{}.
\newblock \showarticletitle{SAUC: Sparsity-Aware Uncertainty Calibration for Spatiotemporal Prediction with Graph Neural Networks}. In \bibinfo{booktitle}{\emph{Temporal Graph Learning Workshop@ NeurIPS 2023}}.
\newblock


\bibitem[Zhuang et~al\mbox{.}(2022)]%
        {zhuang2022uncertainty}
\bibfield{author}{\bibinfo{person}{Dingyi Zhuang}, \bibinfo{person}{Shenhao Wang}, \bibinfo{person}{Haris Koutsopoulos}, {and} \bibinfo{person}{Jinhua Zhao}.} \bibinfo{year}{2022}\natexlab{}.
\newblock \showarticletitle{Uncertainty quantification of sparse travel demand prediction with spatial-temporal graph neural networks}. In \bibinfo{booktitle}{\emph{Proceedings of the 28th ACM SIGKDD Conference on Knowledge Discovery and Data Mining}}. \bibinfo{pages}{4639--4647}.
\newblock


\end{thebibliography}

\end{document}